\pdfoutput=1

\documentclass[11pt]{article}

\usepackage[]{naacl2021}

\usepackage{booktabs}
\usepackage{times}
\usepackage{latexsym}

\usepackage[T1]{fontenc}

\usepackage[utf8]{inputenc}

\usepackage{microtype}
\usepackage{gb4e}
\noautomath
\usepackage{enumitem}
\usepackage{hyperref} 
\usepackage{listings}
\usepackage{xcolor}
\usepackage[]{algorithm2e}
\usepackage{multirow}

\SetKwInOut{Steps}{number of training steps}

\colorlet{punct}{red!60!black}
\definecolor{background}{HTML}{EEEEEE}
\definecolor{delim}{RGB}{20,105,176}
\colorlet{numb}{magenta!60!black}
\lstdefinelanguage{json}{
    basicstyle=\normalfont\ttfamily,
    numbers=left,
    numberstyle=\scriptsize,
    stepnumber=1,
    numbersep=8pt,
    showstringspaces=false,
    breaklines=true,
    literate=
     *{0}{{{\color{numb}0}}}{1}
      {1}{{{\color{numb}1}}}{1}
      {2}{{{\color{numb}2}}}{1}
      {3}{{{\color{numb}3}}}{1}
      {4}{{{\color{numb}4}}}{1}
      {5}{{{\color{numb}5}}}{1}
      {6}{{{\color{numb}6}}}{1}
      {7}{{{\color{numb}7}}}{1}
      {8}{{{\color{numb}8}}}{1}
      {9}{{{\color{numb}9}}}{1}
      {:}{{{\color{punct}{:}}}}{1}
      {,}{{{\color{punct}{,}}}}{1}
      {\{}{{{\color{delim}{\{}}}}{1}
      {\}}{{{\color{delim}{\}}}}}{1}
      {[}{{{\color{delim}{[}}}}{1}
      {]}{{{\color{delim}{]}}}}{1},
}

\usepackage[colorinlistoftodos,prependcaption,textsize=tiny]{todonotes}
\usepackage{xspace}
\usepackage{mathtools}
\DeclarePairedDelimiterX{\infdivx}[2]{(}{)}{%
  #1\;\delimsize|\delimsize|\;#2%
}
\newcommand{\kld}[2]{\ensuremath{D_{KL}\infdivx{#1}{#2}}\xspace}

\definecolor{orangycolor}{HTML}{CC4125}
\definecolor{greensea}{HTML}{16A085}

\usepackage{xargs} 
\newcommandx{\arian}[2][1=]{\todo[linecolor=black,backgroundcolor=teal!10,bordercolor=teal,#1]{Arian: #2}\xspace}
\newcommandx{\alex}[2][1=]{\todo[linecolor=blue,backgroundcolor=blue!10,bordercolor=blue,#1]{ALEX: #2}\xspace}
\newcommandx{\siva}[2][1=]{\todo[linecolor=red,backgroundcolor=red!10,bordercolor=red,#1]{SR: #2}\xspace}
\newcommandx{\dima}[2][1=]{\todo[linecolor=red,backgroundcolor=green!10,bordercolor=red,#1]{DB: #2}\xspace}
\newcommandx{\devon}[2][1=]{\todo[linecolor=yellow,backgroundcolor=yellow!10,bordercolor=red,#1]{DH: #2}\xspace}
\newcommandx{\aaron}[2][1=]{\todo[linecolor=black,backgroundcolor=gray!10,bordercolor=black,#1]{Aaron: #2}\xspace}

\title{Understanding by Understanding Not: \\ Modeling Negation in Language Models}

\author{Arian Hosseini \\
  Mila/Universit\'e de Montr\'eal \\
  Montr\'eal, Canada \\
  \texttt{arian.hosseini9@gmail.com} \\
  \And
  Siva Reddy \\
  Mila/McGill University \\
  Montr\'eal, Canada \\
  \And
  Dzmitry Bahdanau \\
  Element AI\\
  a ServiceNow Company \\
  Montr\'eal, Canada \\
  \AND
  R Devon Hjelm \\
  Mila/Universit\'e de Montr\'eal \\
  and Microsoft Research \\
  Montr\'eal, Canada \\
  \And
  Alessandro Sordoni \\
  Microsoft Research \\
  Montr\'eal, Canada \\
  \And
  Aaron Courville  \\
  Mila/Universit\'e de Montr\'eal \\
  Montr\'eal, Canada
  }

\begin{document}
\maketitle
\begin{abstract}
Negation is a core construction in natural language. Despite being very successful on many tasks, state-of-the-art pre-trained language models often handle negation incorrectly. To improve language models in this regard, we propose to augment the language modeling objective with an unlikelihood objective that is based on negated generic sentences from a raw text corpus. By training BERT with the resulting combined objective we reduce the mean top~1 error rate to 4\% on the negated LAMA dataset. We also see some improvements on the negated NLI benchmarks. 
\end{abstract}

\section{Introduction}

Negation is an important property in many language understanding tasks, such as sentiment analysis, question answering, knowledge base completion and natural language inference \citep{kassner2019negated,DBLP:conf/coling/NaikRSRN18}. While Pre-trained Language Models (PLMs) such as BERT pushed the state-of-the-art on these tasks \citep{DBLP:conf/naacl/DevlinCLT19,DBLP:conf/emnlp/PetroniRRLBWM19}, they fail dramatically on instances that require understanding negation.

\begin{figure}[ht]
    \centering
    \includegraphics[width=0.9\linewidth]{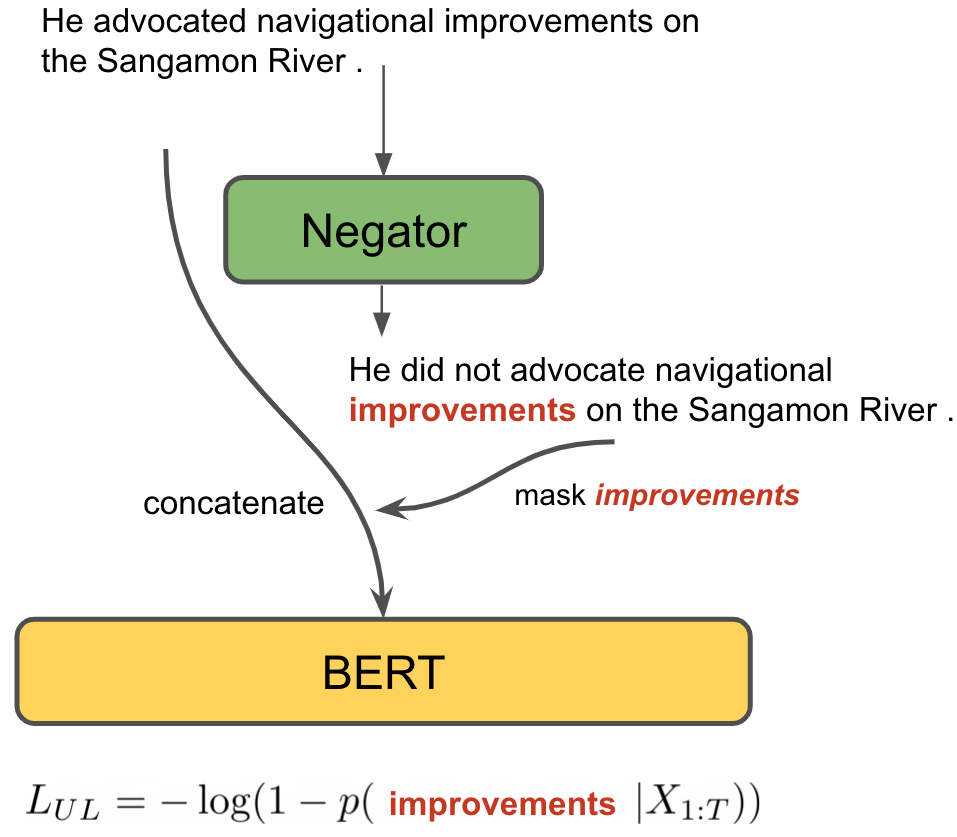}
    \caption{An overview of the unlikelihood objective. A generic sentence is negated using our data augmentation method and an unlikelihood token is chosen and replaced with [MASK]. This new sentence is concatenated with the original sentence and fed into the model. The unlikelihood loss is computed using $p(\textit{\textcolor{orangycolor}{improvements}})$ from the language modeling head of BERT.}
    \label{fig:ul}
\end{figure}

\citet{kassner2019negated} show that current PLMs cannot correctly distinguish between the negated and non-negated forms of fill-in-the-blank tests. For instance, when asked to predict the [MASK] token in sentences such as ``\emph{The capital of Cuba is [MASK]}''  and ``\emph{The capital of Cuba is not [MASK]}'', BERT often generate the same answer ``\emph{Havana}'', indicating that it may not be appropriately modeling the distribution of negative sentences. Additional evidence is given by the fact that, when fine-tuned on natural language inference tasks, PLMs tend to mis-classify examples which contain \textit{not} or \textit{no} as contradiction when the true label is neutral or entailment~\citep{DBLP:conf/coling/NaikRSRN18}.
Recently,~\citet{DBLP:conf/emnlp/HossainKDKWB20} proposed new natural language inference test sets to specifically target the model's understanding of negation and show that current state-of-the-art models perform poorly on these test sets.

In this work, we investigate whether we can alleviate the modeling bias of PLMs on negated sentences. Our approach is composed of two core contributions: i) a syntactic data augmentation scheme to automatically generate negated sentences; ii) a new training paradigm, dubbed~\textit{unlikelihood training with reference} (Fig.~\ref{fig:ul}), based on the recently proposed unlikelihood training~\cite{DBLP:conf/iclr/WelleckKRDCW20}.

At first, we generate a large number of negated sentences by negating sentences mined from an openly available text corpus (Wikipedia). Our sentence negator uses the dependency parse of the sentence, part of speech tags, and morphological features of each word in the sentence and deterministically negates the sentence. 
Given a negated version of a sentence, we replace its object with the [MASK] token and use unlikelihood training to make the object unlikely under the PLM distribution (e.g. we minimize the probability of ``\emph{improvements}'' as depicted in Fig.~\ref{fig:ul}).
Importantly, in order to ensure that the negated sentence is~\emph{factually} false, we use the positive sentence as context (i.e., as a reference) for the unlikelihood prediction task. Concretely, we provide the concatenation of the positive and the masked negated sentence as input to the PLM.
Our method can be thought of a type data augmentation, which has be shown to be effective at improving robustness across many tasks in language, such as text classification \citep{DBLP:conf/emnlp/WeiZ19}, natural language inference \cite{DBLP:journals/corr/abs-2004-11999,  DBLP:conf/acl/McCoyPL19} and semantic parsing \cite{DBLP:journals/corr/abs-1904-09545}. 

For our negation experiments, we fine-tune pre-trained BERT with our new objective and a knowledge distillation objective.
We test our model on the \textit{negated LAMA dataset}~\citep{kassner2019negated}, which is the negated version of knowledge probing dataset LAMA, introduced in~\citet{DBLP:conf/emnlp/PetroniRRLBWM19}. Our model achieves a mean error rate of 4\% (a improvement of 5 points) on the negated LAMA dataset while maintaining the performance on the original LAMA dataset without any direct training on the negated LAMA sentences.
We also fine-tune BERT on RTE ~\citep{DBLP:conf/mlcw/DaganGM05, rte2, giampiccolo-etal-2007-third, Bentivogli09b.:the}, SNLI ~\citep{bowman-etal-2015-large} and MNLI ~\citep{N18-1101} tasks and achieve better results on the language inference benchmark including negation from \cite{DBLP:conf/emnlp/HossainKDKWB20} \footnote{code: \href{https://github.com/arianhosseini/negation-learning}{https://github.com/arianhosseini/negation-learning}}.



\section{Related Work}
\label{sec:related}
Pre-trained language models have shown impressive results across many tasks, such as question answering \cite{DBLP:journals/corr/abs-1901-08634} and natural language inference \cite{DBLP:journals/corr/abs-1907-11692}. These models are also known to encode factual and common-sense knowledge \cite{radford2019language, DBLP:conf/emnlp/PetroniRRLBWM19,bosselut-etal-2019-comet}.
Despite these abilities, \citet{kassner2019negated} found that these models fail at understanding negation through analysing negated factual statements.

Extensive literature looks at the linguistic knowledge learned by language models \cite{DBLP:conf/acl/McCoyPL19, jumelet-hupkes-2018-language, gulordava-etal-2018-colorless, marvin-linzen-2018-targeted, bertpipeline, DBLP:journals/corr/abs-1901-03438, DBLP:journals/corr/abs-1912-13283}. 
Recent work has also studied the shortcomings in negation scope detection \cite{jumelet-hupkes-2018-language, fancellu-etal-2016-neural, fancellu-etal-2017-detecting, DBLP:conf/conll/MoranteD09, li-lu-2018-learning, zhao-bethard-2020-berts, attentionnegationchen} and focus detection \cite{DBLP:conf/emnlp/ShenZHZZA19, DBLP:conf/acl/ZouZZ14, DBLP:conf/emnlp/ZouZZ15, hossain-etal-2020-predicting}.
\citet{DBLP:conf/coling/NaikRSRN18} and \citet{DBLP:conf/acl/McCoyPL19} systematically study the linguistic abilities of these models using NLI, and show that these models rely on erroneous syntactic heuristics.
Our work is in this spirit for negations.

\citet{DBLP:journals/corr/abs-2004-02451} propose taking advantage of negative examples and unlikelihood in the training of language models to increase their syntactic abilities. 
Similarly, \citet{DBLP:journals/corr/abs-2004-11999} show the effectiveness of syntactic data augmentation in the case of robustness in NLI.
Neither of these works focus on negations.

\section{Syntactic Negation Augmentation}
\begin{table*}[ht!]
\centering
\begin{tabular}{l | c c c c}
 \textbf{Model} & {\textbf{SQuAD}} & {\textbf{ConceptNet}} & {\textbf{T-REx}} & {\textbf{Google-RE}}\\
  \hline
BERT  & 13.53 & \textbf{15.65} & 29.10 & 10.24\\
BERT + KL & 13.64 & 15.64 & \textbf{29.28} & 10.27\\
BERTNOT & \textbf{13.97} & 15.49 & 29.25  & \textbf{10.31}\\
\hline
\end{tabular}
\caption{\label{tab:lama} Mean precision at $k=1$ (\textit{p @ 1}) for original LAMA queries (higher is better) of pre-trained BERT, BERT trained with distillation objective, and BERT with unlikelihood and distillation objectives (BERTNOT, sec \ref{sec:ul}). The scores are averaged across 3 runs. }
\end{table*}

\begin{table*}[ht!]
\centering
\begin{tabular}{l | c c c c}
 \textbf{Model} & {\textbf{SQuAD}} & {\textbf{ConceptNet}} & {\textbf{T-REx}} & {\textbf{Google-RE}}\\
  \hline
BERT  & 8.61 & 2.24 & 21.42 & 3.76\\
BERT + KL & 4.97 &  1.19 & 21.77 & 3.99\\
BERTNOT & \textbf{2.10} & \textbf{0.73} & \textbf{11.86} & \textbf{1.10} \\
\hline
\end{tabular}
\caption{\label{tab:neglama} Mean top 1 error rate for negated LAMA queries (lower is better) of pre-trained BERT, BERT trained with distillation objective, and BERT with unlikelihood and distillation objectives (BERTNOT, sec \ref{sec:ul}). The scores are averaged across 3 runs. }
\end{table*}
We generate the negated versions of sentences using a syntactic augmentation method. The method gets as input the dependency parse of the sentence, POS tags and morphological information of each word and negates the sentence using a set of rules.
Each rule has a dependency tree regular expression pattern (Semgrex; ~\citealt{DBLP:conf/acl/ChambersCGHKMMR07}).
We use Semgrex patterns to identify different syntactic templates, and then transform the sentence based on a list of actions defined in the rule. 
These actions can be \emph{move}, \emph{replace}, \emph{insert} and \emph{lemmatize}. 
The unlikelihood token which will be discussed later is also chosen using Semgrex patterns (see Appendix \ref{sec:appendix_neg_aug} for some examples).

We use Stanza \cite{DBLP:journals/corr/abs-2003-07082} to get the dependency parse of the sentences, parts of speech tags, lemma, and morphological features of the words. We filter out sentences with more than 20 words. 

To test the coverage of our Semgrex patterns, we randomly sampled 930 sentences from Wikipedia. Only 31 of them did not match any of our Semgrex patterns (See table \ref{tab:ruleset} in Appendix \ref{app:neg_details_rules} for the number of matches for each rule in our rule set for these 930 sentences). In addition, to get a better sense of the correctness of our method,  100 random sentences (from Wikipedia) were negated and reviewed by a native English speaker. The precision for these negations is $94.00\%$. Table \ref{tab:misnegateds} in Appendix \ref{app:neg_details_rules} shows examples of original and negated sentences.

\section{Unlikelihood Training With Reference}
\subsection{Reference setup}
Applying unlikelihood to a word in any random sentence is problematic, unless the sentence is a factual statement (e.g. unlikelihood on \textit{improvements} in ``He did not advocate navigational \textbf{improvements} on the Sangamon River.'' in Fig \ref{fig:ul} is problematic as this sentence is not grounded in reality). 
Moreover, using solely factual sentences limits the application of this method.\footnote{We did try to apply unlikelihood without any context or reference, but as expected it performed poorly for both LAMA and negated LAMA. See appendix \ref{app:noref}.}
To be able to use any generic (not necessarily factual) sentence and pick an unlikelihood token in it, there needs to be some sort of grounding or context.
In this setup, each training example is of the form $<$\textit{sentence \textbf{A}}, \textit{sentence \textbf{B}}$>$ where \textit{sentence \textbf{A}} is the reference for \textit{sentence \textbf{B}}, and provides the grounding or context for it. 

\subsection{Unlikelihood and knowledge distillation}
\label{sec:ul}
The unlikelihood loss has recently been proposed by \citet{DBLP:conf/iclr/WelleckKRDCW20} to mitigate the problem of repetition in neural text generation. \citet{DBLP:journals/corr/abs-2004-02451} also adopted this loss to penalize the desirability of an incorrect token in a sentence. We adopt this method to penalize the likelihood of a token in \textit{sentence \textbf{B}} that makes this sentence contradictory with the reference \textit{sentence \textbf{A}}.
\begin{exe}
\small
\ex
\begin{enumerate}[label=\Alph*]
\label{ex:UL}
    \item Humans have a rational soul.
    \item Humans do not have a rational \textbf{soul}.
\end{enumerate}
\end{exe}
In the example \ref{ex:UL}, assuming that \textit{sentence \textbf{A}} is true, we want the model to avoid assigning ``soul'' in \textit{sentence \textbf{B}} a high probability. To this end, the probability of the unlikelihood token $x_{u} = \textrm{``soul''}$ is penalized with the unlikelihood loss $L_{UL}$ as:
\begin{equation}
\label{eq:ul}
L_{UL}(x_{u}) = - \log (1 - p(x_{u} | x_{1:T})),
\end{equation}
where $x_{1:T}$ is the whole input sequence (\textit{sentence \textbf{A}} concatenated with \textit{sentence \textbf{B}} which is the negated version of \textit{sentence \textbf{A}} as illustrated in Fig \ref{fig:ul}). To have a balanced augmentation data set, we also include examples where \textit{sentence \textbf{B}} is the copy of \textit{sentence \textbf{A}} and therefore not contradictory with it. In this context, we want the model to perform as it was untouched (before any fine-tuning). The KL divergence knowledge distillation loss is used for these examples on the same token: 

\begin{exe}
\small
\ex
\begin{enumerate}[label=\Alph*]
\label{ex:LL}
    \item Humans have a rational soul.
    \item Humans have a rational \textbf{[MASK]}.
\end{enumerate}
\end{exe}
The loss $L_{KL}$ for token $x_{l} = \textrm{``[MASK]''}$ is written as:
\begin{equation}
\label{eq:ll}
 L_{KL}(x_{l}) = \kld{p_{LM}}{p}       
\end{equation}
where $p_{LM}$ is the probability distribution over the vocabulary for the masked token $x_l$ under the LM before any fine-tuning.

\begin{table*}[h!]
\scriptsize
\centering
\renewcommand{\arraystretch}{1.2}
\begin{tabular}{l | l | l }
 \textbf{Query} & \textbf{Top 3 words with log probs from BERT} & \textbf{Top 3 words with log probs from BERTNOT} \\
\midrule
iOS is developed by [MASK]. & {\color{blue} \textbf{Apple}} (-1.8), Google (-2.6), Microsoft (-2.8) & {\color{blue} \textbf{Apple}} (-1.8), Google (-2.5), Microsoft (-2.7) \\
iOS is not developed by [MASK]. & {\bf \color{red} \textit{Apple}} (-1.8), Google (-2.6), Microsoft (-2.8) & Microsoft (-1.8), Google (-2.4), {\bf \color{red} \textit{Apple}} (-3.1)\\
\midrule
The majority of the amazon forest is in [MASK]. & {\color{blue} \textbf{Brazil}} (-2.6), Bolivia (-2.7), Madagascar (-3.1) & {\color{blue} \textbf{Brazil}} (-2.9), Bolivia (-3.1), Mexico (-3.2) \\
The majority of the amazon forest is not in [MASK]. & cultivation (-1.0), {\bf \color{red} \textit{Brazil}} (-3.5), Mexico (-3.5) & cultivation (-2.0), Mexico (-4.1),  France (-4.3)\\
\midrule
Charles Nodier died in [MASK]. & {\color{blue} \textbf{Paris}} (-1.35), Rome (-3.2), office (-3.4)& {\color{blue} \textbf{Paris}} (-1.5), Rome (-3.3), France (-3.6)\\
Charles Nodier did not die in [MASK]. & {\bf \color{red} \textit{Paris}} (-2.4), office (-2.7), France (-2.8) & vain (-3.5), error (-4.0), doubt (-4.5)\\

\midrule
Mac OS is developed by [MASK]. & {\color{blue} \textbf{Apple}}      
 (-1.9), Microsoft (-2.0), Intel (-2.0)& {\color{blue} \textbf{Apple}} (-2.0), Microsoft (-2.0), Intel (-2.1)\\
Mac OS is not developed by [MASK]. & {\bf \color{red} \textit{Apple}} (-1.3), Microsoft (-1.5), IBM (-2.3) &  Microsoft (-2.1), IBM (-2.7), itself (-3.4)\\

\bottomrule
\end{tabular}
\caption{\label{tab:examples} Examples from BERT base before and after training it with the unlikelihood (UL) and KL divergence knowledge distillation (KL) objectives (BERTNOT). Queries are from LAMA and negated LAMA.}
\end{table*}

In our experiments, we use the BERT-base model and further train it with two objectives, the unlikelihood objective (Eq. \ref{eq:ul}) and the knowledge distillation objective (Eq. \ref{eq:ll}). We also use original Wikipedia sentences for the latter to prevent catastrophic forgetting of language modeling. The probability of the unlikelihood token $p(x_{u} | x_{1:T})$ and the distribution for masked token $x_l$
are computed using the language modeling head of the BERT model by replacing $x_u$ and $x_l$ in the input sequences with the [MASK] token. Examples for each objective are sampled uniformly. We will refer to our model as BERTNOT. 

\section{Experiments}

We report our main results on LAMA and Negated LAMA for knowledge base completion. 
The cloze statements from LAMA are facts or commonsense knowledge generated from either subject-relation-object triples (X, rel, Y) or question-answers pairs. The cloze statements for the triples are generated using a template for each relation which includes the placeholders X and Y (e.g. ``X is located in Y''). X is replaced for the subject and Y is replaced with the {[MASK]} token to be predicted by the model. In the question-answer pairs, the answer is replaced with {[MASK]} token. The facts in the LAMA dataset are from multiple sources: 1) Google-RE relations, namely ``place of birth'', ``date of birth'' and ``place of death''; 
2) T-REx, a subset of Wikidata triples with 41 relations \cite{DBLP:conf/lrec/ElSaharVRGHLS18}; 3) ConceptNet with 16 relations \cite{DBLP:conf/acl/LiTTG16}; 4) SQuAD, a subset of 305 context-insensitive questions 
manually rephrased as cloze-style questions \cite{DBLP:conf/emnlp/RajpurkarZLL16}. Negated LAMA was created by manually negating the templates or questions \cite{kassner2019negated}.
Following \citet{DBLP:conf/emnlp/PetroniRRLBWM19} we use mean precision at $k$ (\textit{P @ $k$}) for LAMA. For negated LAMA we report mean top 1 error rate.

\begin{table*}[h]
\centering
\begin{tabular}{l cc cc cc}
\toprule
\textbf{Model} & \multicolumn{2}{c}{\textbf{RTE}} & \multicolumn{2}{c}{\textbf{SNLI}} & \multicolumn{2}{c}{\textbf{MNLI}}\\
& dev & w/neg & dev & w/neg & dev & w/neg \\ 
\midrule
BERT & $\textbf{70.04}_{\pm 1.57}$ & $65.47_{\pm 3.63}$ & $\textbf{89.47}_{ \pm 0.18}$ & $44.18_{ \pm 0.67}$ & $82.95_{ \pm 0.18}$ & $60.62_{ \pm 1.32}$ \\
BERTNOT & $69.68_{ \pm 1.88}$  & $\textbf{74.47}_{ \pm 0.29}$ & $89.00_{ \pm 0.10}$ & $\textbf{45.96}_{ \pm 0.41}$ & $\textbf{84.31}_{ \pm 2.29}$ & $\textbf{60.89}_{ \pm 0.31}$\\

\bottomrule
\end{tabular}
\caption{\label{tab:mnli} Accuracies on original development splits (dev) and new splits containing negation from \citet{DBLP:conf/emnlp/HossainKDKWB20} (w/neg) for RTE, SNLI and MNLI (matched genres) tasks. Results are averaged across 3 runs.}
\end{table*}

\subsection{Knowledge Base Completion}
As discussed in section \ref{sec:ul}, we train a pre-trained BERT base cased model for 5 epochs, with 20k examples for each objective, a maximum sequence length of 128 and a learning rate of 1e-5. To see the effects of the unlikelihood objective more clearly, we also train a pre-trained BERT base cased model with only the KL knowledge distillation objective with the same data and hyper-parameters.

Tables \ref{tab:lama} and \ref{tab:neglama} respectively show the mean precision at rank 1 (averaged over all the relations) for LAMA, and mean top 1 error rate for negated LAMA queries.\footnote{Baseline scores differ slightly from \citet{DBLP:conf/emnlp/PetroniRRLBWM19}. We were unable to get the same results with their code.} The mean error rate on the negated LAMA queries decreases to below 4\% while the results on original LAMA stay the same. These results are achieved without any direct training on LAMA queries (negated or non-negated). Table \ref{tab:examples} shows the top 3 predicted words for a pre-trained BERT model and the model trained with our method. Pre-trained BERT seems to ignore negation and mostly predict based on the subject of the query, but the prediction probability in the negated queries seems to be generally lower. Our method is as good as the vanilla model (BERT) on original queries.
For the negated queries, our model predictions are far-superior than the vanilla model. We also tried out method on BERT-large. See appendix \ref{app:noref} for results and discussion.

\subsection{Natural Language Inference}
We fine-tune our model with a language inference objective on RTE, SNLI and MNLI tasks. Table~\ref{tab:mnli} shows the accuracies on the original development splits and the new splits from \citet{DBLP:conf/emnlp/HossainKDKWB20} containing negation for each task. We used the hyper-parameters from \citet{DBLP:conf/emnlp/HossainKDKWB20} to fine-tune all of our models.
\begin{table*}[tb!]
\centering
\small
\renewcommand{\arraystretch}{1.2}
\begin{tabular}{c p{6.2cm}| p{6.2cm} | c c c }
 &\textbf{Premise} & \textbf{Hypothesis} & {\textbf{T}} & {\textbf{B}} & {\textbf{BN}}\\
  \hline
1&It does not use the first day of the first month of the Lunar Year as the start of the Chinese New Year. & The Chinese New Year's Day falls on the first day of the first month of the Lunar Year. & N & E & N\\
\hline
2&The prosecutor told the court that the incident had caused "distress" to one of the children. & The prosecutor did not tell the court that "distress" in one of the children is associated with the incident. & N & E & N\\ 
\hline
3&Green cards are not becoming more difficult to obtain. & Green card is now difficult to receive. & N & E & N\\
\hline

4&Moog's synthesiser, which bears his name, revolutionised music from the 1960s onwards, and was used by bands like The Beatles and The Doors. & Moog's instruments were not used by The Beatles and The Doors among others. & N & N & E \\
\hline
5&The board of Marks \& Spencer will not take another look at Philip Green's increased takeover offer. & Philip Green does not try to take over Marks \& Spencer. & E & E & N\\
\hline
6&Albert Sabin developed an oral, attenuated (live) vaccine, which, with Salk's discovery, did not bring polio under control. & Polio is not under control in the world. & E & E & N\\

\hline
\end{tabular}
\caption{\label{tab:negrte}  Examples from the new split from \citet{DBLP:conf/emnlp/HossainKDKWB20} containing negation for RTE. {T}, {B} and {BN} denote true label, BERT's prediction and BERTNOT's prediction respectively. E and N are used for entailment and not entailment labels.}
\end{table*}

Our model achieves superior results on RTE (low-resource setting) and slightly better accuracies on SNLI and MNLI (high-resource setting) on all the new splits containing negation, while keeping roughly the same scores on the original dev splits.
We conjecture that fine-tuning on large-amounts of data (SNLI and MNLI) may have resulted in catastrophic forgetting of the negation knowledge, decreasing the gap between BERT and BERTNOT.
We tried to alleviate the catastrophic forgetting by mixing in some unlikelihood training and knowledge distillation along the NLI training, but that did not help. You can see these results for MNLI in appendix \ref{app:mnliulll}. We leave further exploration of better fine-tuning objectives while preserving the pretrained knowledge for future work.

Table \ref{tab:negrte} shows some of the examples of the new RTE split containing negation from \citet{DBLP:conf/emnlp/HossainKDKWB20}, along with the predictions from BERT and BERTNOT. Examples 4 and 6 show the failure cases of BERTNOT. As it can be seen, for the fifth example, the true label is incorrect, but BERTNOT predicts the correct label for this pair of premise and hypothesis.

\section{Conclusion}
In this work, we propose a combination of the unlikelihood objective with a reference based setup for input sentences to model negation. This allows us to utilize generic sentences, and negate them with our data augmentation method to be used as examples for the unlikelihood objective. Our method notably improves the error rate on the negated LAMA dataset while keeping the same performance on the original LAMA queries.

We also test our method on the original development sets and new splits containing negation from \citet{DBLP:conf/emnlp/HossainKDKWB20} of RTE, SNLI and MNLI tasks.
We see large improvements on the negated splits in low-resource setting (RTE) and slight improvements in high-resource setting (SNLI and MNLI), while also maintaining similar results as BERT on original splits.

\newpage
\bibliographystyle{acl_natbib}
\bibliography{anthology,emnlp2020}


\newpage
\appendix
\onecolumn
\section{Training details}
\label{app:hp}
Here are the hyper-parameters used in our fine-tunings.

\begin{table*}[h]
    \centering
    \small
    \begin{tabular}{c c c c c }
    \hline
        Task & Epochs & Batch Size & Learning Rate & Weight Decay \\
        \hline
        Unlikelihood training & 5 & 32 & 1e-5 & N\textbackslash A\\
        RTE & 50 & 32 & 2e-5 & N\textbackslash A\\
        SNLI & 3 & 32 & 1e-5 & 0.1\\
        MNLI & 3 & 32 & 2e-5 & N\textbackslash A \\
        \bottomrule
    \end{tabular}
    \caption{Hyper-parameters}
    \label{tab:hyperparams}
\end{table*}

Algorithm \ref{alg:ul} shows the details of further training the BERT base cased model with the unlikelihood and knowledge distillation objectives. 
\begin{algorithm}[h]
\small
    \SetAlgoLined
    \Steps{$\mathbf{T}$}
    \For{$i\leftarrow 1$ \KwTo $T$}{
        $L_{UL} \leftarrow$ compute unlikelihood loss with contradictory $<$\textit{sentence \textbf{A}}, \textit{sentence \textbf{B}}$>$ pairs\;
        $L_{KL} \leftarrow$ compute knowledge distillation loss with non-contradictory $<$\textit{sentence \textbf{A}}, \textit{sentence \textbf{B}}$>$ pairs\;
        $g_1 \leftarrow $ compute gradient of $\gamma L_{UL} + (1-\gamma) L_{KL}$\;
        update the parameters with $g_1$\;
        $L_{KL} \leftarrow$ compute knowledge distillation loss with sentences from Wikipedia\;
        $g_2 \leftarrow $ compute gradient of $L_{KL}$\;
        update the parameters with $g_2$\;
    }
    
    \caption{Details of the training procedure of BERTNOT. The unlikelihood loss and knowledge distillation loss are first computed with the $<$\textit{sentence \textbf{A}}, \textit{sentence \textbf{B}}$>$ inputs. These inputs are contradictory for the UL loss, and non-contradictory for knowledge distillation (sec \ref{sec:ul}). We use $\gamma=0.4$ in our experiments to sum these losses and compute the gradient $g_1$. Then, we compute the knowledge distillation loss for inputs sampled from Wikipedia. These inputs do not have our reference based format. The parameters are updated again using the gradient from this knowledge distillation loss ($g_2$).}
    \label{alg:ul}
\end{algorithm}

\pagebreak
\section{Examples of negated sentences}
\label{app:neg_details_rules}
Here are some examples and details of our syntactic negation method.
\begin{table*}[h]
    \centering
    \begin{tabular}{c p{5.3cm}|p{5.3cm}|c}
        &\textbf{Original} & \textbf{Negated} & \textbf{Unlikelihood Token}\\
        \hline
        1&That tournament helped demonstrate the high caliber of play in women's soccer. & That tournament did not help demonstrate the high caliber of play in women's soccer. & tournament\\ 
        \hline
        2&The attributes of this vector (length and direction) characterize the rotation at that point. & The attributes of this vector (length and direction) do not characterize the rotation at that point. & rotation\\
        \hline
        3&This was broadcast live on Norway\textquotesingle s main national TV carrier NRK. & This was not broadcast live on Norway\textquotesingle s main national TV carrier NRK. & Norway\\
        \hline
        4&The latter may occur implicitly through the use of a construct like DEFVAR or DEFPARAMETER. & The latter may not occur implicitly through the use of a construct like DEFVAR or DEFPARAMETER. & latter \\
        \hline
        5&When Arjuna was fighting Karna, the latter\textquotesingle s chariot\textquotesingle s wheels sank into the ground. & When Arjuna was fighting Karna, the latter\textquotesingle s chariot\textquotesingle s wheels did not sank into the ground. & wheels \\
        \hline
        6&It also prohibits or restricts the use of certain accounts held at financial institutions. & It also does not prohibit or restricts the use of certain accounts held at financial institutions. & use\\ 
    \end{tabular}
    \caption{Examples of original and negated sentences with the chosen unlikelihood token. Examples 5 and 6 are incorrect negations since \textit{sank} in example 5 and \textit{restricts} in example 6 are incorrect word forms in the negated context.}
    \label{tab:misnegateds}
\end{table*}

\begin{table*}[h!]
    \centering
    
    \begin{tabular}{c|c }
    \hline
        Rule Name & \# of Sentences Matched \\
        \hline
        simple past & 315 \\
        simple present & 295 \\
        Imperative & 93 \\
        present with auxiliary verb & 37 \\
        past perfect & 35 \\
        copula statements & 34 \\
        present with modal & 24 \\
        already negated with not & 14 \\
        NPI words (anywhere, anyone, etc) & 5 \\
        negative words (no, nobody, etc) & 4 \\
        other & 13\\
        \hline
    \end{tabular}
    \caption{Number of matches for each rule in our rule set over 930 sentences used to analyze the syntactic negation.}
    \label{tab:ruleset}
    
\end{table*}

\pagebreak
\section{Example rules for transforming a sentence into its negation}
\label{sec:appendix_neg_aug}
\lstset{escapeinside={<@}{@>}}
\begin{table*}[ht!]
\centering
\small
\begin{tabular}{p{4cm} | p{6cm} | p{4cm}}
 \textbf{Original Sentence} & \textbf{Rule} & \textbf{Negated Sentence}\\
 \hline 
 \textcolor{greensea}{Nowhere} in his confession \textcolor{blue}{did} he \textcolor{red}{mention} the Monteagle \textcolor{purple}{letter}. & 
 \begin{lstlisting}[language=json, numbers=none, basicstyle=\scriptsize]
{
    "name": "aux before subj",
    "pattern": "{$;tag:/VB.*/}=<@\textcolor{red}{\textbf{A}}@> >/advmod|cc/ {word:/never|nobody|no|nothing|nowhere|neither|Never|Nobody|No|Nothing|Nowhere|Neither/}=<@\textcolor{greensea}{\textbf{npiword}}@> >/aux.*/ ({}= <@\textcolor{blue}{\textbf{B}}@> $++ {}=subject) >/nsubj.*/ {}=subject ?>obj {tag:/NN.*/}=<@\textcolor{purple}{\textbf{object}}@>",
    "actions": [
      {
        "type": "move",
        "to_move": "<@\textcolor{blue}{\textbf{B}}@>",
        "anchor": "<@\textcolor{red}{\textbf{A}}@>",
        "position": "before"
      },
      {
        "type": "replace",
        "token": "",
        "to_replace": "<@\textcolor{greensea}{\textbf{npiword}}@>"
      }
    ]
  }
\end{lstlisting} & in his confession he \textcolor{blue}{did} \textcolor{red}{mention} the Monteagle \textcolor{purple}{letter}. \\ \hline
 Many fonts then \textcolor{red}{made} the right \textcolor{purple}{leg} vertical. & 
 \begin{lstlisting}[language=json, numbers=none, basicstyle=\scriptsize]
  {
    "name": "simple past",
    "pattern": "{$;cpos:/.*Tense=Past.*/}=<@\textcolor{red}{\textbf{A}}@> >/nsubj|csubj/=E {}=subject ?>obj {tag:/NN.*/}=<@\textcolor{purple}{\textbf{object}}@>",
    "actions": [
      {
        "type": "insert",
        "token": "did",
        "rel": "AUX",
        "anchor": "<@\textcolor{red}{\textbf{A}}@>",
        "position": "before"
      },
      {
        "type": "insert",
        "token": "not",
        "rel": "ADV",
        "anchor": "<@\textcolor{red}{\textbf{A}}@>",
        "position": "before"
      },
      {
        "type": "lemmatize"
      }
    ]
  }
 \end{lstlisting}
 & Many fonts then did not \textcolor{red}{make} the right \textcolor{purple}{leg} vertical.\\
 

\end{tabular}
\caption{\label{tab:neg_aug} Examples of how the syntactic negation augmentation method works. For the first sentence, the matched rule has two actions, \textit{move} and \textit{replace}. The \textit{move} action has moved the token \textcolor{blue}{\textbf{B} = did} before token \textcolor{red}{\textbf{A} = mention}. The \textit{replace} action has replaced \textcolor{greensea}{\textbf{npiword} = Nowhere} with an empty token, which means removing this token. The token \textcolor{purple}{\textbf{object} = letter} is chosen as the unlikelihood token in this sentence.\\
In the second sentence, the matched rule has three actions, two \textit{insert}s and one \textit{lemmatize} action. The \textit{insert} actions, add the tokens ``did not'' before \textcolor{red}{\textbf{A} = made}, and the token \textcolor{red}{\textbf{A} = made} is replaced with its lemma by the \textit{lemmatize} action. The token \textcolor{purple}{\textbf{object} = leg} is chosen as the unlikelihood token in the negated sentence.}
\end{table*}


\section{Mixing negation unlikelihood training and knowledge distillation with NLI training}
\label{app:mnliulll}
In order to reduce the catastrophic forgetting behavior of the model during NLI training, we added the unlikelihood, knowledge distillation and MLM objectives to the original NLI classification objective and trained the model with the same hyper-parameters for the MNLI task. We also trained one version with only the original NLI classification objective and the MLM objective. As the results in table \ref{tab:mnliulll} show, this method did not improve the scores for development split and the new split containing negation from \citet{DBLP:conf/emnlp/HossainKDKWB20} for MNLI.

\begin{table*}[h]
\footnotesize
\centering
\begin{tabular}{l cc}
\toprule
\textbf{Model} & \multicolumn{2}{c}{\textbf{MNLI}}\\
& dev & w/neg \\ 
\midrule
BERTNOT + UL + KL + MLM + NLI obj & 81.17 & 60.20 \\ 
BERTNOT + MLM + NLI obj & 81.42 & 62.00\\
\bottomrule
\end{tabular}
\caption{\label{tab:mnliulll} Accuracies on original development split (dev) and new split containing negation from \citet{DBLP:conf/emnlp/HossainKDKWB20} (w/neg) for MNLI (matched genres) task.}
\end{table*}

\section{Supplementary Results}
\label{app:noref}

\begin{table*}[ht!]
\centering
\small
\begin{tabular}{l c | c c c c}
 \textbf{Model} & {\textbf{lr}} & {\textbf{SQuAD}} & {\textbf{ConceptNet}} & {\textbf{T-REx}} & {\textbf{Google-RE}}\\
  \hline
BERTNOT without reference setup & 1e-5 & 13.86 & 15.65 & 29.54 & 10.29\\
BERT-large & 1e-5 & 16.83 & 19.26 & 30.76 & 10.93\\
BERTNOT-large & 1e-5 & 14.19 & 19.14 & 32.09 & 11.02\\
BERTNOT-large & 5e-5  & 15.18 & 16.97 & 30.71 & 10.62\\
BERTNOT-large & 1e-4  & 11.55 & 13.58 & 28.41 & 9.25\\
\hline
\end{tabular}
\caption{\label{tab:lamanoref} Mean precision at $k=1$ (\textit{p @ 1}) for original LAMA queries (higher is better) of BERT with unlikelihood and distillation objectives without references for sentences, BERT-large, and BERT-large with unlikelihood and distillation objectives with different learning rates.}
\end{table*}

\begin{table*}[ht!]
\centering
\small
\begin{tabular}{l c | c c c c}
 \textbf{Model} & {\textbf{lr}} & {\textbf{SQuAD}} & {\textbf{ConceptNet}} & {\textbf{T-REx}} & {\textbf{Google-RE}}\\
  \hline
BERTNOT without reference setup & 1e-5  & 5.96 & 1.34 & 21.54 & 3.73\\
BERT-large & 1e-5 & 7.95 & 1.67 & 22.97 & 4.13\\
BERTNOT-large & 1e-5  & 8.28 & 1.87 & 23.49 & 4.22\\
BERTNOT-large & 5e-5  & 8.28 & 2.20 & 24.05 & 4.09\\
BERTNOT-large & 1e-4  & 4.97 & 1.47 & 20.86 & 3.60\\
\hline
\end{tabular}
\caption{\label{tab:neglamanoref} Mean top 1 error rate for negated LAMA queries (lower is better) of BERT with unlikelihood and distillation objectives without references for sentences, BERT-large, and BERT-large with unlikelihood and distillation objectives with different learning rates.}
\end{table*}

As the results in table \ref{tab:neglamanoref} show, pre-trained BERT-large performs worse than pre-trained BERT-base on negated LAMA queries. We decreased the batch-size to be able to fine-tune BERT-large. As the scores for negated LAMA queries from table \ref{tab:neglamanoref} show, fine-tuning BERT-large with our method using the same or slightly larger learning rate does not improve the results. We observe a decrease in the mean top 1 error rates for negated LAMA queries when we use a larger learning rate ($1e-5$), but this also hinders the performance of the model on the original LAMA queries (table \ref{tab:lamanoref}). This requires some hyper-parameter tuning and further investigation.

\end{document}